\documentclass[letterpaper, 10 pt, conference]{ieeeconf}  % Comment this line out if you need a4paper

\IEEEoverridecommandlockouts                              % This command is only needed if 
\usepackage{graphicx} % for pdf, bitmapped graphics files
\usepackage{amsmath} % assumes amsmath package installed
\usepackage{amssymb}  % assumes amsmath package installed
\usepackage{subfig}
\usepackage{placeins}
\usepackage{afterpage}
\usepackage{fewerfloatpages}
\usepackage{color}
\usepackage{tikz}

\newcommand\copyrighttext{%
	\footnotesize \textcopyright \the\year{} IEEE. Personal use of this material is permitted. Permission from IEEE must be obtained for all other uses, in any current or future media, including reprinting/republishing this material for advertising or promotional purposes, creating new collective works, for resale or redistribution to servers or lists, or reuse of any copyrighted component of this work in other works.\\
	This is the accepted manuscript and the published version can be found at:https://ieeexplore.ieee.org/document/11020953. \\
	DOI: 10.1109/RoboSoft63089.2025.11020953}
\newcommand\copyrightnotice{%
	\begin{tikzpicture}[remember picture, overlay]
	\node[anchor=north, yshift=-4pt] at (current page.north) {\parbox{\dimexpr\textwidth-\fboxsep-\fboxrule\relax}{\textcolor{red}{\copyrighttext}}};
	\end{tikzpicture}%
}
\title{\LARGE \bf
Model Predictive Control for a Soft Robotic Finger with Stochastic Behavior based on Fokker-Planck Equation
}

\author{Sumitaka Honji$^{1}$ and Takahiro Wada$^{1}$% <-this % stops a space
\thanks{*This work was not supported by any organization}% <-this % stops a space
\thanks{$^{1}$Sumitaka Honji and Takahiro Wada are with the Division of Information Science, Graduate School of Science and Technology,
        Nara Institute of Science and Technology, Ikoma, Nara, Japan.
        {\tt\small honji.sumitaka@naist.ac.jp} and {\tt\small t.wada@is.naist.jp}}%
}

\begin{document}

\maketitle

\copyrightnotice

%%%%%%%%%%%%%%%%%%%%%%%%%%%%%%%%%%%%%%%%%%%%%%%%%%%%%%%%%%%%%%%%%%%%%%%%%%%%%%%%
\begin{abstract}

% original
% Though the flexibility of soft robots brings a lot of benefits such as adaptability and safety, it can cause uncertain and varied motion.
% This could be a problem especially when a soft robot is actuated by open-loop controllers which are major control methods because there is no feedback system.
% Model-based control is one of the solutions to this problem and the stochastic point of view is important to deal with the uncertainty of soft robots.
% Here we show the implementation of a stochastic-based controller, which is called the Fokker-Plank Equation-based Model Predictive Control for a soft robotic finger.
% Through 2 case studies in numerical simulation, the characteristics of this control method are revealed.
% GPT version
The inherent flexibility of soft robots offers numerous advantages, such as enhanced adaptability and improved safety.
However, this flexibility can also introduce challenges regarding highly uncertain and nonlinear motion.
These challenges become particularly problematic when using open-loop control methods, which lack a feedback mechanism and are commonly employed in soft robot control.
Though one potential solution is model-based control, typical deterministic models struggle with uncertainty as mentioned above.
% Therefore, incorporating a stochastic approach is essential to address the uncertainties associated with soft robots.
The idea is to use the Fokker-Planck Equation (FPE), a master equation of a stochastic process, to control not the state of soft robots but the probabilistic distribution.
In this study, we propose and implement a stochastic-based control strategy, termed FPE-based Model Predictive Control (FPE-MPC), for a soft robotic finger.
Two numerical simulation case studies examine the performance and characteristics of this control method, revealing its efficacy in managing the uncertainty inherent in soft robotic systems.

\end{abstract}

%%%%%%%%%%%%%%%%%%%%%%%%%%%%%%%%%%%%%%%%%%%%%%%%%%%%%%%%%%%%%%%%%%%%%%%%%%%%%%%%
\section{INTRODUCTION}
% original
% Soft robots are expected to perform in many industrial fields such as agricultural picking or marine investigation \cite{sinatra2019ultragentle}, and in human-robot interaction.
% The flexible property of soft robots serves an important role: the softness can realize the high adaptive behavior of soft robots, which enables them to pick fragile or unshaped objects; interact with physical safety, etc.
% GPT version
Soft robots are increasingly being utilized in various industrial sectors, including agriculture (e.g., fruit picking) \cite{elfferich2022soft} and marine exploration \cite{sinatra2019ultragentle}, as well as in human-robot interaction scenarios.
A key advantage of soft robots lies in their inherent flexibility, which allows for highly adaptive behavior.
This softness enables them to handle fragile or irregularly shaped objects and to interact safely with humans and their physical environment.
These attributes make soft robots a promising alternative to traditional rigid robots, particularly in tasks that require delicate manipulation or enhanced safety measures.

% original
% Though soft robots can provide many interesting features that traditional hard robots cannot do, their functionality mainly depends on their mechanics or fabrication designs.
% On the other hand, controlling soft robots is still challenging.
% This is because their dynamics is highly nonlinear and difficult to treat mathematically.
% In addition, the difficulty of sensing their state makes it hard to use feedback controllers.
% Therefore a simple controller such as binary control tends to be preferred.
% This type of feedforward control works well thanks to the adaptability of soft robots and the easiness of implementation, however, it realizes mainly the passive behavior and is difficult to function soft them for complicated tasks.
% GPT version
Despite the advantages of soft robots, controlling their movements remains a significant challenge.
Their complex and nonlinear dynamics are difficult to model accurately, making precise control a challenging task \cite{truby2020distributed}.
Furthermore, the soft nature of these robots complicates state sensing, as traditional sensors are often inadequate for capturing the full range of their deformations.
Consequently, simple control methods, such as binary or feedforward control, are commonly used in soft robotics.
While these methods benefit from the ease of implementation and leverage the intrinsic adaptability of soft robots, they typically result in inactive behavior, limiting the robot's ability to perform complex, high-precision tasks.

Modeling soft robots' characteristics is one solution to address the abovementioned concerns.
Many models have been studied from many perspectives such as geometry, lumped approximation, and data-driven methods \cite{armanini2023soft}.
The (piecewise) Constant Curvature model \cite{webster2010design}, Cosserat rod model \cite{renda2014dynamic}, Finite Element model \cite{duriez2013control}, nonlinear dynamics with Koopman operator \cite{bruder2019nonlinear}, and others are representatives.
In addition, these models are used in the framework of model-based control and make it possible to actively manage the behavior of soft robots \cite{della2023model}.
% Various modeling techniques, including Finite Element Models (FEM) \cite{duriez2013control} and Cosserat rod models \cite{boyer2023statics, renda2014dynamic}, have been proposed and applied within control frameworks.
For instance, Duriez implemented interactive control of a soft tripod using a FEM simulator called SOFA \cite{duriez2013control}.
Similarly, Boyer et al. utilized a Cosserat rod model for optimal control of continuum robots \cite{boyer2023statics}.
However, traditional deterministic models cannot deal with the uncertain behavior of soft robots. Regarding this point, stochastic approaches are also gaining attention in soft robotics.
Case et al. demonstrated that the behavior of soft materials, commonly used in soft robot fabrication, varies across different individuals, trials, and environmental conditions \cite{case2015soft}.
This variability underscores the importance of incorporating stochastic models into control frameworks for soft robots.
Related to the above study, Kim et al. extended this concept by modeling a pneumatic soft actuator as a stochastic process using variational inference, enabling state estimation under uncertainty \cite{kim2021probabilistic}.
The author observed similar uncertainty in the behavior of a silicone-based soft robotic finger, leading to the development of a stochastic dynamical model by combining the lumped parameterized approximation with stochastic parameters \cite{honji2023stochastic}.

% original
% The stochastic control approach has been studied in a wide range of research fields.
% As a method that combines model-based and stochastic control, stochastic model predictive controls have been proposed \cite{mesbah2016stochastic}.
% These methods are mainly applied to control probabilistic phenomena such as chemical reactions, automobile applications, and so on.
% To deal with the uncertain behavior of soft robots even though there are enough sensors, it is important to control its variability.
% Fokker-Plank Equation-based MPC (FPE-MPC) matches this concept, which controls not the state directly but the probability density function (PDF) \cite{buehler2016lyapunov}.
% FPE is a master equation of a stochastic process and describes the time evolution of a stochastic process in the form of PDF.
% This controller can be conducted in feedforward and therefore is expected to match soft robots.
% GPT version
Stochastic control methods have been widely studied in other fields, and stochastic model predictive control (MPC) has been proposed as a promising approach that integrates both model-based and stochastic control techniques \cite{mesbah2016stochastic}.
These methods have been successfully applied in domains such as chemical process control and automotive systems, where probabilistic phenomena play a critical role.
In the context of soft robotics, controlling variability is essential, even in scenarios where adequate sensing is available.
Therefore, we focus on the Fokker-Planck equation (FPE), which governs the time evolution of a stochastic process in terms of its Probability Density Function (PDF).
In other words, controlling the PDF through FPE means that there is a possibility to control not only the expectation but also other statistical values such as the variance, which will suppress soft robots' uncertainty.
To achieve this, FPE-based Model Predictive Control (FPE-MPC) provides a suitable framework for this, as it focuses on controlling the PDF of the robot's state, rather than the state itself \cite{buehler2016lyapunov}.
% The Fokker-Planck equation, which governs the time evolution of a stochastic process in terms of its PDF, serves as the foundation for this control strategy.
% FPE-MPC, being a feedforward control method, is well-suited for soft robotics, where feedback control is often difficult to implement due to sensing limitations.
They provided that the controlled state could converge within the controlled PDF.

% original
% In this study, we apply FPE-MPC to control a soft robotic finger in feedforward.
% Controlling the PDF that the state follows makes regulating the state within the confidence in the sense of probability.
% Through case studies conducted in simulation, the characteristics of this method are revealed.
% GPT version
In this study, we apply FPE-MPC to the control of a soft robotic finger in a feedforward manner.
By controlling the PDF of the robot’s state, we aim to regulate its behavior within probabilistic confidence bounds.
Through numerical simulations, we evaluate the performance of this method and reveal its potential for managing the uncertainty inherent in soft robotic systems.

\section{CONTROL FRAMEWORK}

\subsection{Fokker-Planck Equation}

First, we consider the following general $n$-dimensional stochastic system with $m$-dimensional input that is written by Ito's stochastic differential equation:
\begin{equation}
    d\boldsymbol{x} = \boldsymbol{f} \left( \boldsymbol{x, u}, t \right) dt + \boldsymbol{\sigma} \left( \boldsymbol{x}, t \right) d \boldsymbol{W} \left( t \right), \label{eq:1}
\end{equation}
here $\boldsymbol{x} \in \mathbb{R}^n, \boldsymbol{u} \in \mathbb{R}^m$ are the state and the input of the system, while $\boldsymbol{f}\left(\boldsymbol{x}, \boldsymbol{u}, t\right)$ is its advection term.
Also, $\boldsymbol{\sigma} \left( \boldsymbol{x}, t \right) \in \mathbb{R}^{n \times l}$ expresses the diffusion coefficient and $\boldsymbol{W} \left( t \right) \in \mathbb{R}^{l}$ is the $l$-dimensional standard Wiener process.
The FPE is derived as a master equation for this system, describing the stochastic process transition as below:
%
% \begin{align}
%     \frac{d p \left( \boldsymbol{x} \right)}{dt} &= \frac{\partial}{\partial \boldsymbol{x}} \left\{ \boldsymbol{B \left( x \right)} + \frac{\partial}{\partial \boldsymbol{x}} \boldsymbol{C \left( x \right)} \right\} \quad \in \mathbb{R}, \label{eq:2} \\
%     \boldsymbol{B \left( x \right)} &= - \boldsymbol{f \left( x, u \right)} p \left( \boldsymbol{x} \right) \quad \in \mathbb{R}^n, \notag \\
%     \boldsymbol{C \left( x \right)} &= \frac{1}{2} \boldsymbol{\sigma} ^2 \left( \boldsymbol{x} \right) p \left( \boldsymbol{x} \right) \quad \in \mathbb{R}^{n\times n}, \notag
% \end{align}
\begin{align}
    \frac{\partial p \left( \boldsymbol{x} \right)}{\partial t} &= \frac{1}{2} \sum_{i,j=1}^n \frac{\partial^2}{\partial x_i \partial x_j} \left\{ a_{i,j} \left( \boldsymbol{x},t \right) p \left( \boldsymbol{x}, t \right) \right\} \notag \\
    &\quad - \sum_{i=1}^n \frac{\partial}{\partial x_i} \left\{ f_i \left( \boldsymbol{x, u}, t \right) p \left(\boldsymbol{x} ,t \right)\right\}, \label{eq:2} \\
    \boldsymbol{a} \left( \boldsymbol{x}, t \right) &= \boldsymbol{\sigma} \cdot \boldsymbol{\sigma}^\top \quad \in \mathbb{R}^{n \times n}, \notag
\end{align}
$p \left( \boldsymbol{x} \right)$ is the PDF which the state $\boldsymbol{x}$ follows, and $\boldsymbol{a} \left( \boldsymbol{x}, t\right)$ works as the space and time dependent diffusion coefficient.
This means that if the shape of the PDF can be controlled, we can control statistic values such as the expectation and the variance.
Let us remark that \eqref{eq:2} is the general advection-diffusion equation and we no longer need to consider the stochastic term directly.

\subsection{Discretization}

Eq. \eqref{eq:2} is a partial differential equation and is difficult to solve theoretically.
To obtain numerical solutions, many methods such as the finite volume method or upwind method have been proposed for accurate and stable solutions.
In addition, because PDF is always positive, a method that can guarantee the positiveness of a solution is necessary.
For this reason, the Chang-Cooper scheme which has 2nd-order accuracy for the time and space domain and guarantees positiveness is adopted \cite{chang1970practical}.
Here is a brief explanation and please see \cite{annunziato2013fokker,mohammadi2015analysis} for more detail.
From here, the 2-dimensional case ($\boldsymbol{x} = \left[ \begin{array}{cc}
    x_1 & x_2
\end{array} \right]^\top$) which is the same as our implementation in Section \ref{sec:simulation} is provided for simplicity.
In this case, the value of the PDF $p \left( \boldsymbol{x} \right)$ at $x_1 = ih, x_2 = jh$ is represented as $p_{i,j}$ with spatial resolution $h$.

First, for the time domain, the 2nd order forward differential is used as,
\begin{equation}
    \frac{\partial p_{i,j}}{\partial t}\left[k+1\right] \approx \frac{3 p_{i,j}\left[k+1\right] - 4 p_{i,j}\left[k\right] + p_{i,j}\left[k-1\right]}{2 dt}, \label{eq:3}
\end{equation}
Here $a \left[ k \right]$ means the value of a function or a variable $a$ at time step $k$.
$dt$ is the time step resolution.

Second, the implicit 1st order Euler differential is utilized for the space domain.
Eq. \eqref{eq:2} can be written as the conservative flux form.
In fact, by deforming \eqref{eq:2}, this can be seen as flux form and the flux is defined as follows:
% The flux is defined as follows:
%
\begin{align}
    \frac{\partial p}{\partial t} &= \sum_{i=1}^n \frac{\partial}{\partial x_i} \left\{ \left( \frac{1}{2} \sum_{j=1}^n \frac{\partial a_{i,j}}{\partial x_j} - f_i \right) p \right. \notag \\
    &\quad + \left. \frac{1}{2} \sum_{j=1}^n a_{i,j} \frac{\partial p}{\partial x_j} \right\}, \notag \\
    & = \sum_{i=1}^n \frac{\partial}{\partial x_i} P^i, \label{eq:4} \\
    P^i &= \frac{1}{2} \sum_{j=1}^n a_{i,j}\frac{\partial p}{\partial x_j} + \left( \frac{1}{2} \sum_{j=1}^n \frac{\partial a_{i,j}}{\partial x_j} - f_i \right) p, \notag \\
    &= \sum_{j=1}^n C^{i,j} \frac{\partial p}{\partial x_j} + B^i p. \label{eq:5}
    % B^i &= \frac{1}{2} \sum_{j=1}^n \frac{\partial a_{i,j}}{\partial x_j} - f_i, \notag \\
    % C^{i,j} &= \frac{1}{2} a_{i,j}. \notag
\end{align}
%
% and using this definition leads to the stochastic flux equation as,
% \begin{equation}
%     \frac{\partial p_{i,j}}{\partial t} = \nabla \cdot P_{i,j}, \label{eq:4}
% \end{equation}
% %
% where $P_{i,j}$ is a stochastic flux at $x_1 = ih, x_2 = jh$.
% From here, we show the 2-dimensional case for simplicity as:
We omit the variables for simplicity.
Then, from \eqref{eq:4} the spatial derivative of the stochastic flux is discretized in the following manners:
\begin{align}
    \frac{\partial p_{i,j}}{\partial t} &= \frac{1}{h} \left( P_{i+\frac{1}{2},j}^1\left[k \right] - P_{i-\frac{1}{2},j}^1\left[k \right] \right. \notag \\
    &\quad\left. + P_{i,j+\frac{1}{2}}^2\left[k \right] - P_{i,j-\frac{1}{2}}^2\left[k \right] \right), \label{eq:6} \\
% \end{align}
%
% where $P_{i,j}$ means the $i,j$ element of stochastic flux $\boldsymbol{P}$, $i,j$ are discretized spatial position, and $h$ is the spatial resolution.
% And the discretized flux at time step $k$ is calculated as,
%
% \begin{align}
    P_{i+\frac{1}{2},j}^i \left[k \right] &= \left( \left( 1 - \delta_i \right) B_{i+\frac{1}{2}, j}^i \left[k \right] + \frac{1}{h} C_{i+\frac{1}{2}, j}^i \left[k \right] \right) \notag \\ 
    &\quad \times p_{i+\frac{1}{2}, j}\left[k+1 \right] \notag \\
    &\quad - \left( \frac{1}{h} C_{i+\frac{1}{2}, j}^i \left[k \right] - \delta_i B_{i+\frac{1}{2}, j}^i \left[k \right] \right) p_{i, j} \left[k+1 \right], \label{eq:7} \\
    \delta_i &= \frac{1}{w_i} - \frac{1}{\exp\left(w_i\right) - 1}, \notag \\
    w_i &= h \frac{B_{i+\frac{1}{2},j}^i \left[k \right]}{C_{i+\frac{1}{2},j}^i \left[k \right]}, \notag
\end{align}
$\delta_i$ comes from the condition that the flux $P$ equals 0 at the edge of considered space domain.
Combining \eqref{eq:3} and \eqref{eq:6} with \eqref{eq:7}, PDF values at $k+1$ can be obtained by solving the algebraic equation.

\subsection{Model Predictive Control}

The nonlinear Model Predictive Control (MPC) framework can be applied to \eqref{eq:2} to obtain the optimal input to control the shape of PDF.
The objective function is designed as,
\begin{equation}
    J \left[ k \right] = \sum_{i,j=1}^n \left( p_{i,j} \left[ k \right] - ^\mathrm{ref}p_{i,j} \left[ k \right] \right)^2, \label{eq:8}
\end{equation}
with the reference PDF $^\mathrm{ref} p \left( \boldsymbol{x} \right)$.
The optimal input $\boldsymbol{u}^*$ can be obtained to minimize the objective function.

\section{SOFT FINGER MODEL}

FPE-MPC is used to control the soft robotic finger.
Authors have proposed the dynamic model of a kind of soft finger whose viscoelastic parameters are expressed as stochastic variables \cite{honji2023stochastic}.
This model aims to represent the joint creep behavior by the 3-element viscoelastic combination and the variable and uncertain motion of a soft finger by stochastic variables.
Here is a brief explanation of this model.

\begin{figure*}[tb!]
    \centering
    \includegraphics[width=0.8\linewidth]{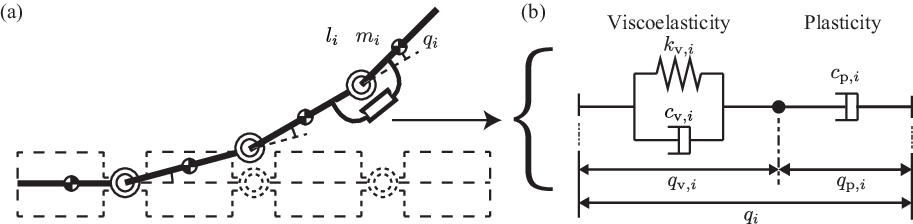}
    \caption{Overall finger model construction. (a) linked manipulator approximation, and (b) joint viscoelastic 3-element combination are shown.}
    \label{fig:1}
\end{figure*}
% \afterpage{\clearpage}

\subsection{Dynamics}
This model consists of 2 structures: the whole body is approximated by a linked manipulator and the joints are expressed by a joint viscoelastic combination as shown in Fig. \ref{fig:1}.
When we approximate a soft finger as $n_j$ joint manipulator, the model becomes as:
\begin{equation}
    \left\{
    \begin{array}{c}
        \boldsymbol{M \left( q \right) \ddot{q} + h \left( q, \dot{q} \right)} + \boldsymbol{\tau}_\mathrm{joint} = \boldsymbol{P \left( q \right) u} \\
        \boldsymbol{A \dot{q} + B q} = \boldsymbol{C \tau}_\mathrm{joint} + \boldsymbol{D} \int{\boldsymbol{\tau}_\mathrm{joint}}\, dt
    \end{array}
    \right., \label{eq:9}
\end{equation}
where $\boldsymbol{M \left( q \right)} \in \mathbb{R}^{n_j \times n_j}, \boldsymbol{h \left( q, \dot{q} \right)} \in \mathbb{R}^{n_j}$ and $\boldsymbol{\tau}_\mathrm{joint} \in \mathbb{R}^{n_j}$ are the inertia matrix, nonlinear term and joint internal torque respectively.
$\boldsymbol{u} \in \mathbb{R}^m$ expresses the input and $\boldsymbol{P \left( q \right)} \in \mathbb{R}^{n_j \times m}$ does the coupling matrix, and are cable tension and its transformation to control torque respectively.
Furthermore, $\boldsymbol{A, B, C}$ and $\boldsymbol{D} \in \mathbb{R}^{n_j \times n_j}$ are diagonal matrices whose $\left(i, i \right)$ elements are $c_{\mathrm{p}, i} k_{\mathrm{v}, i}, c_{\mathrm{v}, i} c_{\mathrm{p}, i}, c_{\mathrm{v}, i} + c_{\mathrm{p}, i}$ and $k_{\mathrm{v}, i} \left( i = 1, 2, ..., n_j\right)$ respectively.
You can identify the meaning of each parameter from Fig. \ref{fig:1}.
By taking the state variable $\boldsymbol{x} = \left[ \begin{array}{ccc} \boldsymbol{q}^\top & \dot{\boldsymbol{q}}^\top & \boldsymbol{\tau}_\mathrm{joint}^\top \end{array} \right]^\top \in \mathbb{R}^{3n_j}$, the state equation is derived as follows:
\begin{equation}
    \boldsymbol{\dot{x} = f \left( x, u, p \right) = f \left( x, u, \hat{p} \right)} + \frac{\partial \boldsymbol{f}}{\partial \boldsymbol{p}} \left( \boldsymbol{x, u, \hat{p}} \right) \delta \boldsymbol{p} \label{eq:10},
\end{equation}
the RHS is the Taylor 1st order expansion with $\boldsymbol{p}$ around the nominal value $\hat{\boldsymbol{p}}$.
Here $\boldsymbol{p} \in \mathbb{R}^{n_p}$ is a parameter vector that contains the viscoelastic parameters $k_\mathrm{v}, c_\mathrm{v}$ and $c_\mathrm{p}$ and each parameter is expressed as a stochastic variable as explained in the next subsection.
Therefore, $\frac{\partial \boldsymbol{f}}{\partial \boldsymbol{p}}$ and $\delta \boldsymbol{p}$ can be seen $\boldsymbol{\sigma}$ and $d\boldsymbol{W}$ of \eqref{eq:1} and the dynamics is rewritten as the stochastic differential equation.

% \begin{figure*}[tbp]
%     \centering
%     \includegraphics[width=\linewidth]{image/fingermodel.pdf}
%     \caption{Overall finger model construction. (a) linked manipulator approximation, and (b) joint viscoelastic 3-element combination are shown.}
%     \label{fig:1}
% \end{figure*}
% \afterpage{\clearpage}

\subsection{Parameter Uncertainty}
This model assumes that each viscoelastic parameter follows a stochastic distribution.
According to the previous research \cite{honji2024state}, we adopt the log-normal distribution \eqref{eq:11},
\begin{equation}
    PDF_\mathrm{log-normal} \left( x \right) = \frac{1}{x \sqrt{2 \pi \sigma^2}} \exp \left( - \frac{\left( \log{x} - \mu \right)^2}{2 \sigma^2} \right). \label{eq:11}
\end{equation}

The log-normal distribution can be transformed to the normal distribution by stochastic variable transformation $x^\prime = \exp\left(x\right)$ and new stochastic variable $x^\prime$ follows a normal distribution.
By this procedure, shape parameters $\mu, \sigma$ become expectations and variance of transformed normal distribution.

\subsection{2-link 3-tendon soft finger}

\begin{figure}[tb!]
    \centering
    \includegraphics[width=0.8\linewidth]{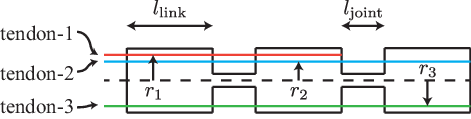}
    \caption{Soft finger structure with 2-link 3-tendon mechanism. $l_\mathrm{link}$ and $l_\mathrm{joint}$ represent the length of each part, and $r_i$ means the distance from the center line (dotted) to tendon-$i$.}
    \label{fig:2l3tfinger}
\end{figure}

In this study, we use the 2-link 3-tendon soft finger model as same as \cite{honji2024state}.
As seen in Fig. \ref{fig:2l3tfinger}.
The actuation tendon-1 passes through only joint-1 and tendon-2 and tendon-3 are managed to actuate joint-1 and -2 for both directions.
Following values are used for each parameter: $l_\mathrm{link} = 30 \times 10^{-3}$, $l_\mathrm{joint} = 15 \times 10^{-3}$ m, $r_i = 8, 5, 8 \times 10^{-3}$ m $\left( i = 1, 2, 3 \right)$, respectively.
Shape parameters for each distribution of a viscoelastic parameter are shown in Table \ref{tab:1} and nominal values become $\exp{\left(\mu\right)}$.

\begin{table}[tb!]
    \centering
    \caption{Shape parameters of log-normal distribution.}
    \label{tab:1}
    \begin{tabular}{c|cccc}
         & \multicolumn{2}{c}{Joint 1} & \multicolumn{2}{c}{Joint 2} \\
        Parameter & $\mu$ & $\sigma$ & $\mu$ & $\sigma$ \\ \hline \hline
        $k_\mathrm{v}$ & -2.81 & 0.0918 & -2.62 & 0.115 \\
        $c_\mathrm{v}$ & -4.20 & 0.648 & -4.27 & 0.492 \\
        $c_\mathrm{p}$ & 2.13 & 0.706 & 2.72 & 0.912
    \end{tabular}
\end{table}

\section{CASE STUDY}\label{sec:simulation}

Simulations were conducted to investigate the performance of FPE-MPC.
The case-1 investigated several reference PDFs with different expectations and the same variance from the initial shape.
The case-2 showed the possibility of changing the variance of PDF when designed.
Both cases considered uncertain viscoelastic parameters as a stochastic source.

\subsection{Preparation}

The FPE that described the transition of PDF for the joint angle was solved, which meant the PDF had a 2-dimensional state value $\boldsymbol{q} = \left[ \begin{array}{cc} q_1 & q_2 \end{array} \right]^\top$.
For both case studies, the same initial conditions \eqref{eq:12} was used as follows:
\begin{align}
    ^\mathrm{ini} p \left( \boldsymbol{q} \right) &= \prod_{i=1}^2 \frac{1}{\sqrt{2 \pi \sigma_{\mathrm{ini}, i}^2}} \exp{\left(- \frac{\left( q_i - \mu_{\mathrm{ini}, i} \right)^2}{2 \sigma_{\mathrm{ini}, i}^2} \right)}, \label{eq:12}
\end{align}
where $\mu_{\mathrm{ini},i} = 0$ and $\sigma_{\mathrm{ini},i} = 0.05$.
Furthermore, the actual initial value of the joint angle was also 0 rad.
The transition of PDF was calculated in the specific region: $-0.15 \leq q_i \leq 0.15$ with spatial resolution $h = 0.15$.
As a result, $21 \times 21 = 441$ points' time evolution was calculated to obtain the optimal input.
Simulation time is 10 sec with a time resolution of 0.1 sec.
In addition, the controller only knows the medians of stochastic parameters $\exp \left( \mu \right)$.
This means that in the controller, the predicted state with nominal viscoelastic parameters is used to determine the control input and the proposed method becomes open-loop completely.

The FPE-MPC was implemented in MATLAB using predefined function \texttt{nlmpc} and all codes were run on a desktop PC with Intel core i7 and 16 GB RAM.
For each prediction step, the controller uses nominal viscoelastic parameters to update the current state, which means that this controller works completely as a feedforward.
The prediction horizon and the control horizon were both set to be 1.
The upper and lower bounds of input are restricted to 5 and 0 N because a cable-driven mechanism is considered in this study.

\subsection{Case-1: difference for reference expectations}

\begin{figure*}[tb!]
    \centering
    \subfloat[][$\mu_{\mathrm{ref},i} = -0.5, 0.5$.]{\includegraphics[width=0.33\linewidth]{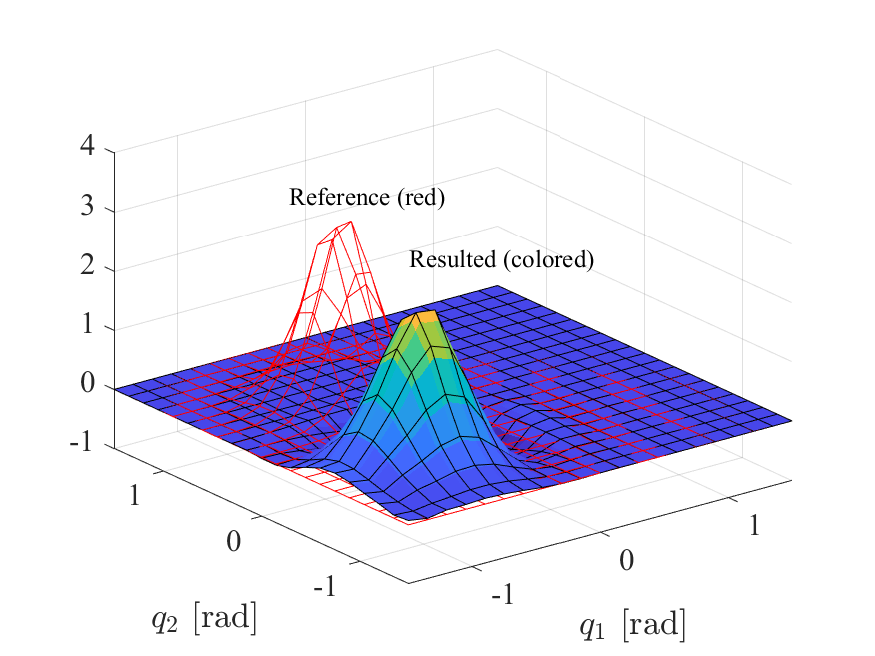}\label{fig:2a}}
    \subfloat[][$\mu_{\mathrm{ref},i} = 0.5, 0.3$.]{\includegraphics[width=0.33\linewidth]{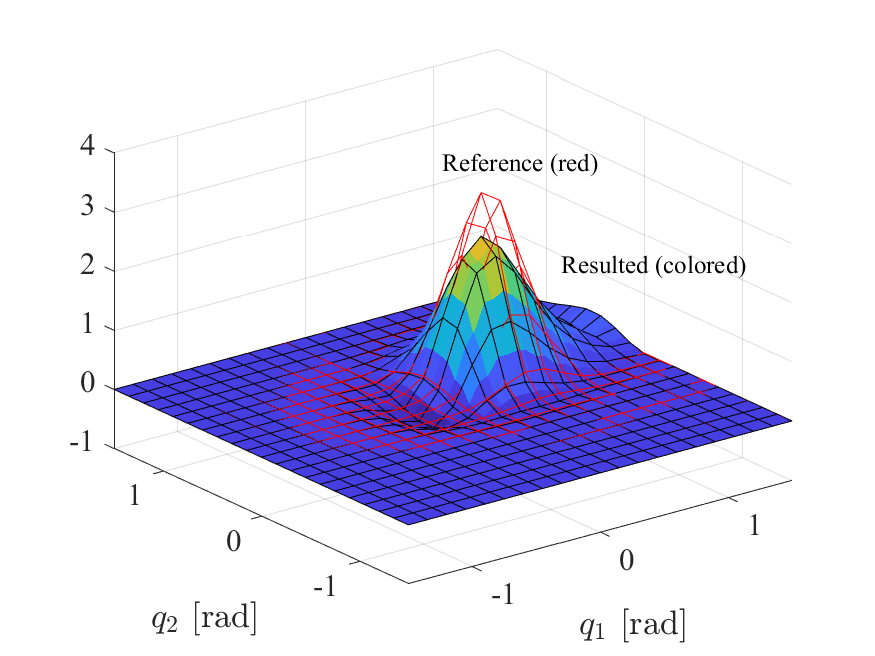}\label{fig:2b}}
    \subfloat[][$\mu_{\mathrm{ref},i} = -0.3, -0.3$.]{\includegraphics[width=0.33\linewidth]{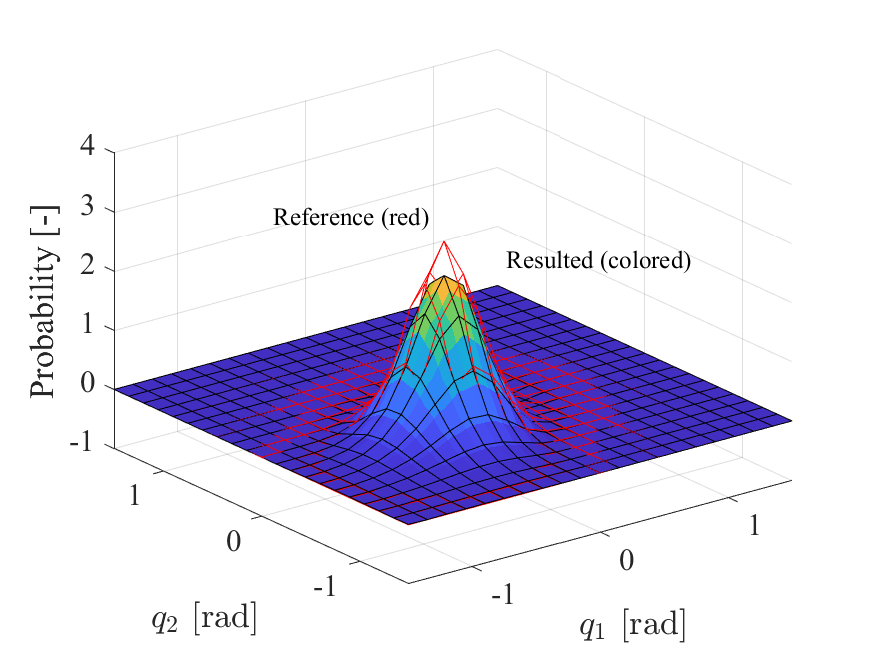}\label{fig:2c}}
    \caption{Resulted (Colored) and reference (red framed) PDFs for expectation sets: \protect\subref{fig:2a} Cannot be controlled. \protect\subref{fig:2b} Controlled around the reference. \protect\subref{fig:2c} Controlled almost close to the reference.}
    \label{fig:2}
\end{figure*}

Firstly, the reference PDF $^\mathrm{ref}p \left(\boldsymbol{q}\right)$ was defined as the 2-dimensional normal distribution \eqref{eq:13} as,
\begin{align}
    ^\mathrm{ref} p \left( \boldsymbol{q} \right) &= \prod_{i=1}^2 \frac{1}{\sqrt{2 \pi \sigma_{\mathrm{ref}, i}^2}} \exp{\left(- \frac{\left( q_i - \mu_{\mathrm{ref}, i} \right)^2}{2 \sigma_{\mathrm{ref}, i}^2} \right)}, \label{eq:13}
\end{align}
We set several combinations of expectations $\mu_{\mathrm{ref},i} = 0, \pm 0.3, \pm 0.5$ but the variance was always same as $\sigma_{\mathrm{ref},i} = \sigma_{\mathrm{ini},i} = 0.05$.
For example, when we choose $\mu_{\mathrm{ref}, i} = 0.5, 0.3$ respectively, the PDFs are set as seen in Fig. \ref{fig:2}.

\begin{figure}
    \centering
    \includegraphics[width=\linewidth]{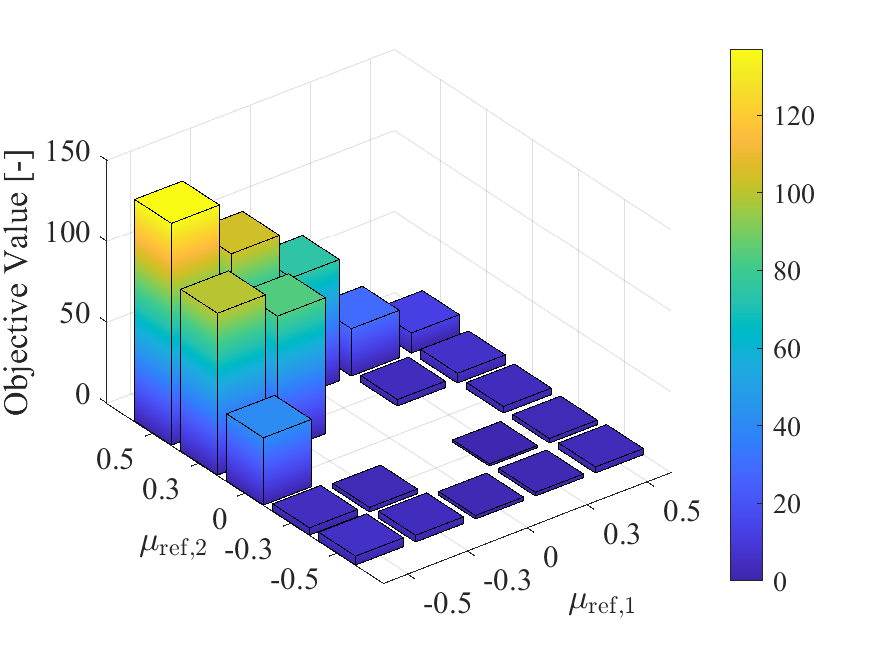}
    \caption{Values of the objective function for each expectation set. $\mu_{\mathrm{ref}, i}$ means the expectation angle for $i$-th joint (seen as desired value). A height bar means the combination of expectations cannot be controlled and vice versa.}
    \label{fig:6}
\end{figure}

% As a result, Table. \ref{tab:2} shows the values of the objective function at $t = 10$ sec.
As a result, Fig. \ref{fig:6} shows the values of the objective function at $t = 10$ sec.
When the bar is high, the combination of reference expectations cannot be controlled, and vice versa.
% Also, some results of controlled PDFs are shown in Fig. \ref{fig:2}, and you can see the graphical images of controlled and uncontrolled PDFs.
The graphical images of controlled and uncontrolled PDFs are shown in Fig. \ref{fig:2}.
These results show that the FPE-MPC can control the shape of the PDF according to its initial and reference values.
The reachable expectation sets come from the reachable area of the soft finger model \eqref{eq:9}.
Because the transition of the PDF is described by the dynamics that FPE is based as seen in \eqref{eq:1}, this implies that designing an actuation mechanism (in this case $\boldsymbol{P \left( x \right)}$ in \eqref{eq:9}) can realize a wide range of PDF reference.

\begin{figure*}[tb!]
    \centering
    \subfloat[][State transition.]{\includegraphics[width=0.33\linewidth]{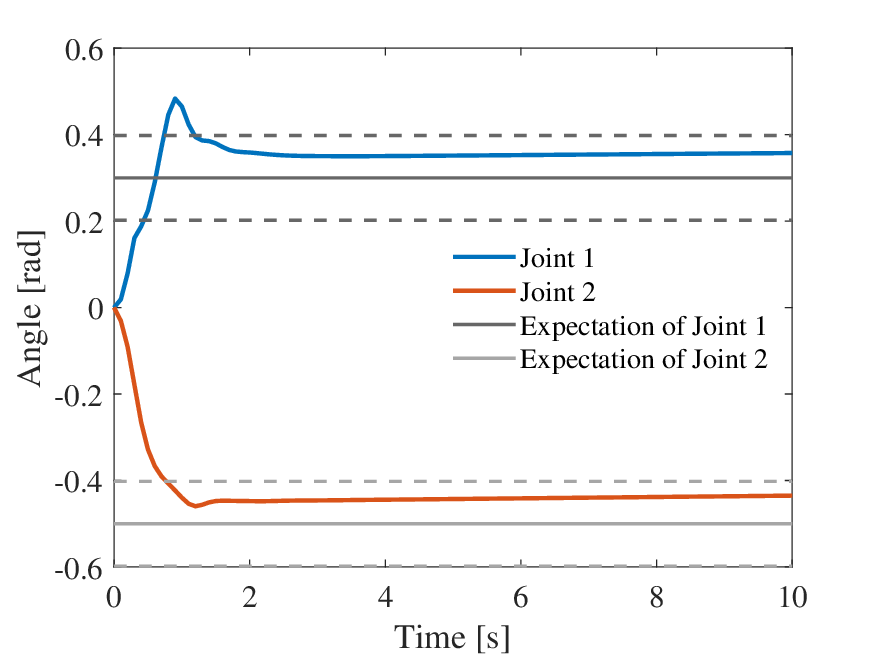}\label{fig:3a}}
    \subfloat[][Calculated input.]{\includegraphics[width=0.33\linewidth]{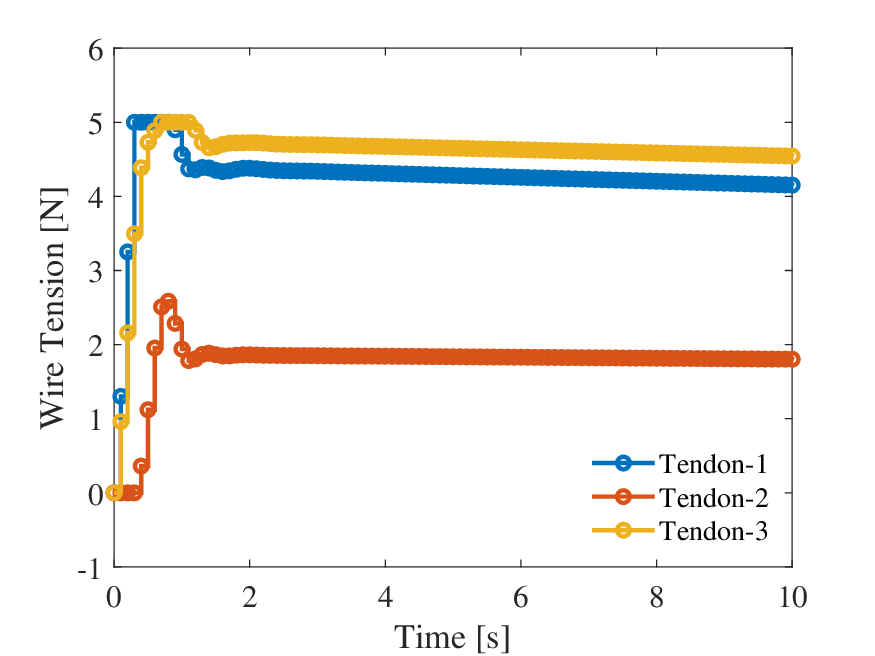}\label{fig:3b}}
    \subfloat[][Objective value transition.]{\includegraphics[width=0.33\linewidth]{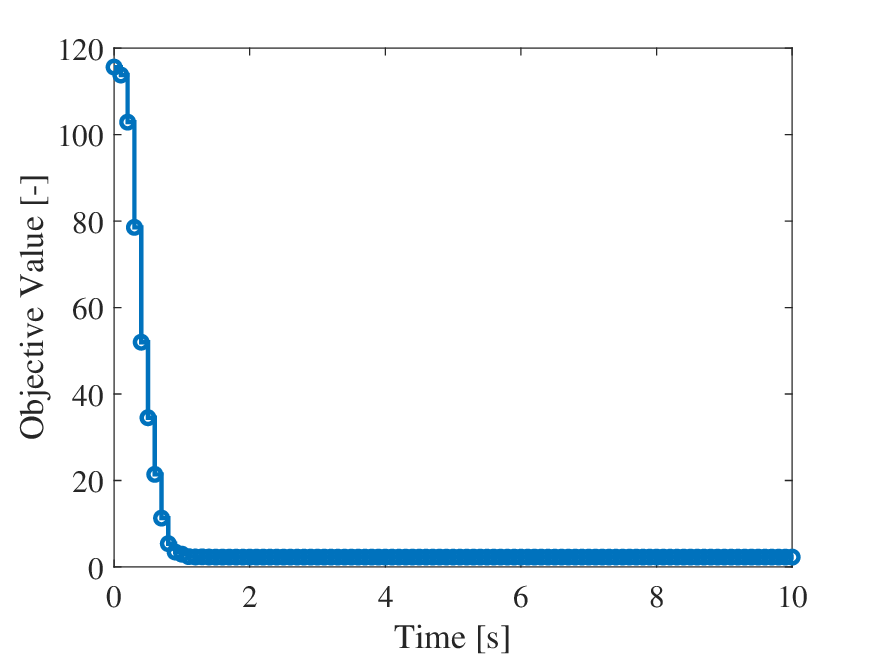}\label{fig:3c}}
    \caption{\protect\subref{fig:3a} Transition of actual joint angle. Solid gray lines are expectations and dotted lines represent 95\% confidence. \protect\subref{fig:3b} Control inputs by FPE-MPC. \protect\subref{fig:3c} Objective value reducing with time steps. ($\mu_{\mathrm{ref}, i} = 0.3, - 0.5, \sigma_{\mathrm{ref}, i} = 0.05, 0.05$)}
    \label{fig:3}
\end{figure*}

Here, we pick up one result that the reference expectations are $\mu_{\mathrm{ref}, i} = 0.3, - 0.5$, as shown in Fig. \ref{fig:3}.
By using the FPE-MPC, the actual state converged around the expectation within 95\% confidence ($\mu_{\mathrm{ref}, i} \pm 1.96 \, \sigma_{\mathrm{ref}, i}$).
Nevertheless, the result strongly depends on the actual viscoelastic parameters that are decided randomly from their distribution in this simulation.
It is revealed that the smaller $c_\mathrm{v}$ takes, the worse the performance becomes.

% \begin{figure}[tbp]
%     \centering
%     \includegraphics[width=\linewidth]{image/PDF_ini_ref.eps}
%     \caption{Initial and reference PDFs in case $\mu_{\mathrm{ref},i} = 0.5, 0.3.$}
%     \label{fig:2}
% \end{figure}
%
% \begin{table}[tb!]
%     \centering
%     \caption{Values of the objective function for each expectation set.}
%     \label{tab:2}
%     \begin{tabular}{cc|ccccc}
%          \multicolumn{2}{c|}{Expectaion set} & \multicolumn{5}{c}{Joint 1} \\
%          & & -0.5 & -0.3 & 0.0 & 0.3 & 0.5 \\ \hline
%          & 0.5 & 137 & 104 & 74.9 & 29.3 & 12.4 \\
%          & 0.3 & 100 & 84.0 & & 4.05 & 6.04 \\
%         Joint 2 & 0.0 & 41.6 & & initial & & 4.04 \\
%          & -0.3 & 4.47 & 2.89 & & 1.35 & 2.82 \\
%          & -0.5 & 5.50 & 4.19 & 2.21 & 2.27 & 3.74
%     \end{tabular}
% \end{table}

\subsection{Case-2: difference for reference variances}

Next, we show the case that the reference PDF has a different variance from the initial one.
This is expected to work in situations where we do not need accuracy for one direction but not for another.

\begin{figure*}[tb!]
    \centering
    \subfloat[][State transition.]{\includegraphics[width=0.33\linewidth]{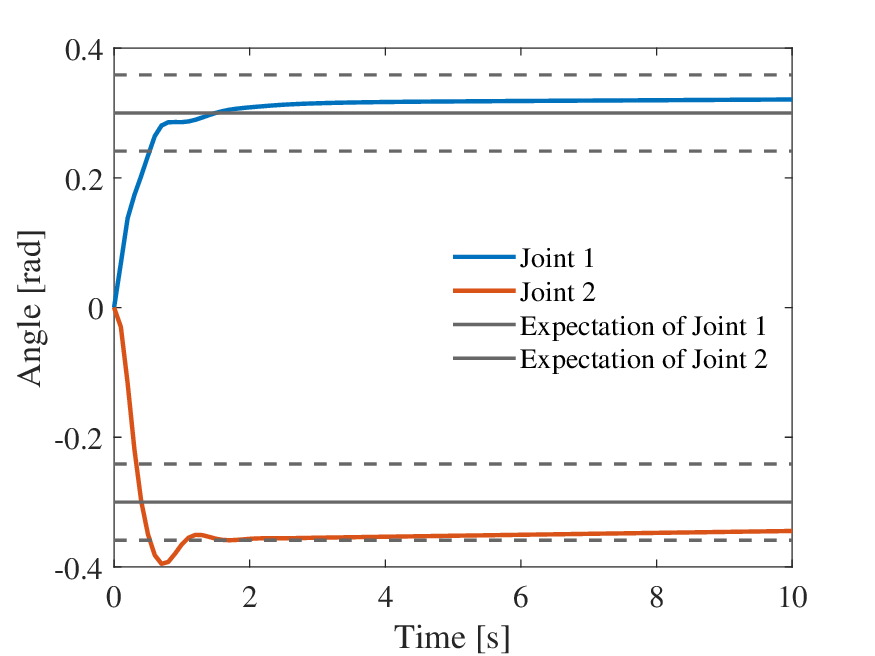}\label{fig:4a}}
    \subfloat[][Calculated input.]{\includegraphics[width=0.33\linewidth]{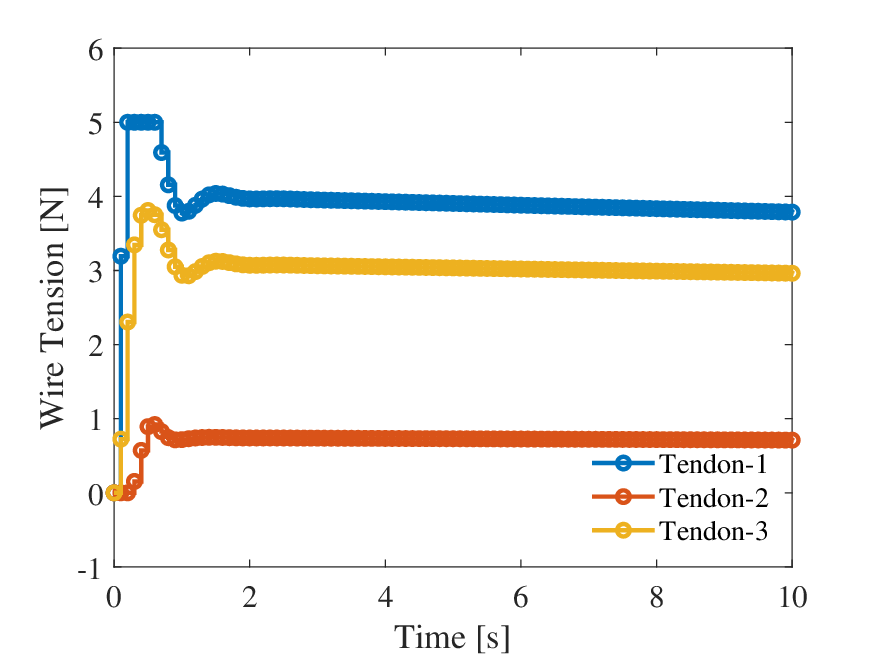}\label{fig:4b}}
    \subfloat[][Objective value transition.]{\includegraphics[width=0.33\linewidth]{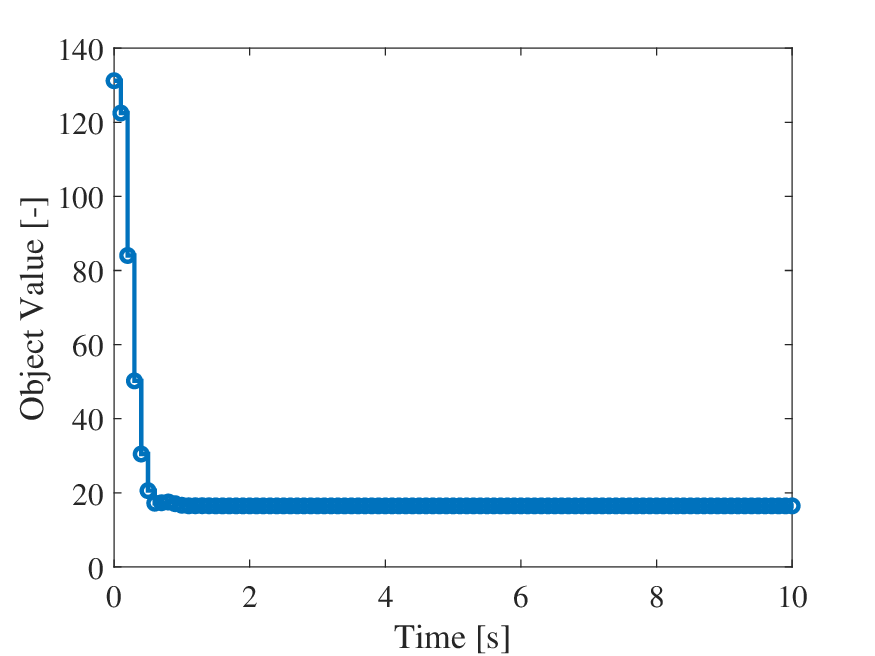}\label{fig:4c}}
    \caption{\protect\subref{fig:3a} Transition of actual joint angle. Solid gray lines are expectations and dotted lines represent 95\% confidence. \protect\subref{fig:3b} Control inputs by FPE-MPC. \protect\subref{fig:3c} Objective value reducing with time steps. ($\mu_{\mathrm{ref}, i} = 0.3, - 0.3, \sigma_{\mathrm{ref}, i} = 0.03, 0.03$)}
    \label{fig:4}
\end{figure*}

\begin{figure}[tb!]
    \centering
    \includegraphics[width=\linewidth]{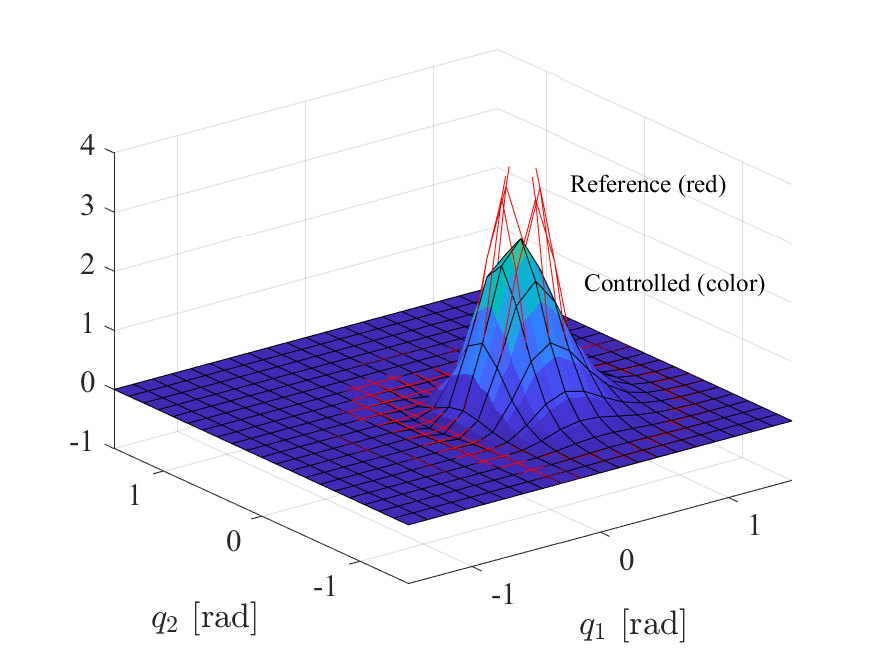}
    \caption{Resulting PDF and its reference when reference variance is set small.}
    \label{fig:5}
\end{figure}

This is the result when we set reference expectations as $\mu_{\mathrm{ref},i} = 0.3, -0.3$ and reference variances as $\sigma_{\mathrm{ref},i} = 0.03, 0.03$ respectively.
Figure \ref{fig:4} shows the controlled joint angle, calculated control input, and the transition of objective value respectively.
However the joint angle looks to converge within the 95\% confidence, the value of the objective function indicates that there is a difference between the controlled PDF and reference PDF as shown in Fig \ref{fig:5}.

We found that forming the PDF in terms of the variance is harder than that of the expectation.
That is, too-narrow PDFs may not be realized by using only the model information.
To achieve this, the diffusion term should be utilized more efficiently to interfere with the diffusion process in the future.

\section{DISCUSSION}

\subsection{Advantages}

In this study, we establish a completely open-loop control method that utilizes the stochasticity of a soft robotic finger.
Through the shape control of PDF, the FPE-MPC controls the actual state to fit within some region in the sense of stochasticity.
% The great advantage of this method is that any sensors are not necessary to control the soft robot, whereas soft robots have an infinite degree of freedom and deform overall their body and this makes it difficult to attach sensors to them.

Though these case studies treated parameter uncertainty, other kinds of uncertainty could be dealt with as long as they are written in the form of $\boldsymbol{\sigma} \left( \boldsymbol{x}, t\right) d \boldsymbol{W} \left(t\right)$ in \eqref{eq:1}.

\subsection{Limitations}

Controlling accuracy is one of the most challenging parts of this method.
% Because there is no feedback loop and even the objective function is calculated in the simulation, it is hard to evaluate the effect for the original dynamics.
Especially, this method requires the initial PDF.
It is considered that control performance will be worse if the initial state is far out of the initial PDF.
However, this implies that the proposed method can be improved by combining with some observing system and feedback its information.
Another point is the computational cost.
As we have mentioned, this method calculates the time evolution of each grid point of the discretized space domain as seen in Fig. \ref{fig:2} or \ref{fig:5} (in this case 441 points) and this will be worse when considering more high dimensional systems, which is called "the curse of dimensionality".
This also brings challenges to the scalability of this method.
When a many DoF system such as hyper-redundant manipulators is considered, it is not reasonable to deal with all possible states, and model reduction techniques should be used.
Nevertheless, many kinds of research have been proposed to implement fast MPC like the C/GMRES method \cite{ohtsuka2004continuation} which helps us improve the performance of FPE-MPC.
Learning-based algorithms such as Physics Informed Neural Networks \cite{raissi2019physics} can also be used.

In summary, though many drawbacks for the FPE-MPC exist, this kind of stochastic control method serves an important role in terms of soft robotics.
For future work, this controller will be combined with a feedback system and implemented as more efficient coding, then applied in the actual situations.
This is also related to assessing what kinds of uncertainty should be treated in actual ones.
In addition, though the proposed method is different from conventional open-loop controllers from the viewpoint of stochasticity, the effectiveness should be analyzed.

\section{CONCLUSIONS}
% original
% In this research, the Fokker-Planck Equation-based Model Predictive Control (FPE-MPC) was implemented in the simulkation and case studies for different reference settings were investigated.
% Depending on the actuation mechanism, FPE-MPC can control the Probability Density Function (PDF) in feedforward and the obtained input can also control the state of simulated soft finger within the 95 \% confidence.
% Otherwise, we found some struggle references, especially when the variance of reference PDF is changed from the initial.

% In the future work, the FPE-MPC can be combined with feedback loop to accurately control the soft robots.
% In addition, practical numerical scheme should be adopted for actual implementation.
% GPT version
In this study, we implemented the Fokker-Planck Equation-based Model Predictive Control (FPE-MPC) in numerical simulations and investigated its performance through case studies with varying reference settings. The results demonstrate that FPE-MPC can control the shape of the Probability Density Function (PDF) in a feedforward manner, allowing the state of the simulated soft robotic finger to be regulated within 95\% confidence. However, challenges were observed in certain scenarios, particularly when there were significant changes in the variance of the reference PDF from the initial conditions.

Future work should focus on enhancing FPE-MPC by integrating it with a feedback loop, which would improve the accuracy and robustness of control for soft robots. Additionally, the development of more efficient numerical schemes is essential to facilitate the practical implementation of this control method in real-world soft robotic systems.

\addtolength{\textheight}{-12cm}   % This command serves to balance the column lengths

%%%%%%%%%%%%%%%%%%%%%%%%%%%%%%%%%%%%%%%%%%%%%%%%%%%%%%%%%%%%%%%%%%%%%%%%%%%%%%%%

%%%%%%%%%%%%%%%%%%%%%%%%%%%%%%%%%%%%%%%%%%%%%%%%%%%%%%%%%%%%%%%%%%%%%%%%%%%%%%%%



% \section*{ACKNOWLEDGMENT}

% The preferred spelling of the word ÒacknowledgmentÓ in America is without an ÒeÓ after the ÒgÓ. Avoid the stilted expression, ÒOne of us (R. B. G.) thanks . . .Ó  Instead, try ÒR. B. G. thanksÓ. Put sponsor acknowledgments in the unnumbered footnote on the first page.

%%%%%%%%%%%%%%%%%%%%%%%%%%%%%%%%%%%%%%%%%%%%%%%%%%%%%%%%%%%%%%%%%%%%%%%%%%%%%%%%

% References are important to the reader; therefore, each citation must be complete and correct. If at all possible, references should be commonly available publications.

\bibliographystyle{IEEEtran}
\bibliography{references}

\end{document}